\pdfoutput=1

\documentclass[11pt]{article}

\usepackage[preprint]{acl}

\usepackage{times}
\usepackage{latexsym}

\usepackage[T1]{fontenc}

\usepackage[utf8]{inputenc}

\usepackage{microtype}

\usepackage{inconsolata}

\usepackage{graphicx}

\usepackage{algorithm}
\usepackage{algpseudocode}
\usepackage{subfigure}

\usepackage{xcolor} 
\usepackage{tikz}   
\usepackage{atbegshi} 

\AtBeginShipout{%
  \AtBeginShipoutUpperLeft{%
    \begin{tikzpicture}[remember picture, overlay]
      \node[anchor=north east, xshift=18.7cm, yshift=-1cm, text=gray!70] at (current page.north west) {[Submitted on 28 Nov 2023 (v1), last revised 17 Jun 2024 (this version, v2)]};
    \end{tikzpicture}%
  }%
}

\title{Large Language Models Suffer From Their Own Output: \\ An Analysis of the Self-Consuming Training Loop}

\author{
 \textbf{Martin Briesch\textsuperscript{1}},
 \textbf{Dominik Sobania\textsuperscript{1}},
 \textbf{Franz Rothlauf\textsuperscript{1}}
\\
 \textsuperscript{1}Johannes Gutenberg University, Mainz, Germany,
\\
 \small{
   \textbf{Correspondence:} \href{mailto:briesch@uni-mainz.de}{briesch@uni-mainz.de}
 }
}

\begin{document}
\maketitle
\begin{abstract}
Large Language Models (LLM) are already widely used to generate content for a variety of online platforms. As we are not able to safely distinguish LLM-generated content from human-produced content, LLM-generated content is used to train the next generation of LLMs, giving rise to a self-consuming training loop. From the image generation domain we know that such a self-consuming training loop reduces both quality and diversity of images finally ending in a model collapse. However, it is unclear whether this alarming effect can also be observed for LLMs. Therefore, we present the first study investigating the self-consuming training loop for LLMs. Further, we propose a novel method based on logic expressions that allows us to unambiguously verify the correctness of LLM-generated content, which is difficult for natural language text. We find that the self-consuming training loop produces correct outputs, however, the output declines in its diversity depending on the proportion of the used generated data. Fresh data can slow down this decline, but not stop it. Given these concerning results, we encourage researchers to study methods to negate this process. 
\end{abstract}

\section{Introduction}
\label{sec:intro}

Transformer-based language models have received much attention in the machine learning community in recent years. Especially large language models (LLM) trained on massive amounts of data from the internet became state of the art in many benchmarks \cite{brown2020language} and specialized conversational LLM applications like ChatGPT\footnote{\href{https://openai.com/blog/chatgpt}{https://openai.com/blog/chatgpt}} have a massive influence on society already. LLMs can be used for multiple tasks ranging from code generation \cite{chen2021evaluating,sobania2022choose,fan2023large} and automated program repair \cite{sobania2023chatgpt} to text summarization \cite{yang2023exploring} and teaching assistance \cite{baidoo2023education}. 

Due to their convincing generated outputs, LLMs can be used to generate a large amount of content that is posted online to coding platforms like GitHub and Stackoverflow\footnote{\href{https://www.microsoft.com/en-us/Investor/events/FY-2023/Morgan-Stanley-TMT-Conference}{https://www.microsoft.com/en-us/Investor/events/FY-2023/Morgan-Stanley-TMT-Conference}}, social media platforms like Reddit\footnote{\href{https://www.vice.com/en/article/jg5qy8/reddit-moderators-brace-for-a-chatgpt-spam-apocalypse}{https://www.vice.com/en/article/jg5qy8/reddit-moderators-brace-for-a-chatgpt-spam-apocalypse}} and other platforms on the internet. Even academic writing is already being influenced by LLM outputs \cite{geng2024chatgpt, liang2024mapping}. Such LLM-generated text is often hard to distinguish from human-generated content \cite{sadasivan2023can} and in turn might unwillingly be used to train the next generation of LLMs. Even paying for human-generated content might not be an option in the future, as workers at paid services like Amazon's Mechanical Turk also use LLMs to produce content \cite{veselovsky2023artificial}.
Consequently, a self-consuming training loop emerges in which future models are trained repeatedly on LLM-generated data from previous generations. 
This process was first observed for generative models in the image domain \cite{martinez2023towards, alemohammad2023self, shumailov2023model, bertrand2023stability}. These studies found that this self-consuming training loop leads to a decline in quality and diversity of generated images, ultimately resulting in a so called model collapse. 
However, it is unclear what happens with LLMs that are trained in such a self-consuming training loop from scratch, as usually done in real-world applications like ChatGPT.

Therefore, we present the first study analyzing the behavior of LLMs trained over many generations in a self-consuming training loop. We conduct experiments on a GPT-style model in different settings and measure both quality and diversity of samples from the trained model at each generation in the self-consuming training loop. The settings differ in the way a dataset is created for each generation (so called data cycles) as well as the proportion of original \textit{real} and LLM-generated \textit{synthetic} data samples. 
To better analyze the behavior of the trained models we conduct our experiments on a dataset consisting of logic expressions. In contrast to natural language, this logic expressions can be evaluated unambiguously. This allows us to analytically and accurately measure correctness and diversity of the generated samples. 

We find that repeatedly training new models with synthetic data from previous models initially improves quality. However, diversity degenerates and the learned distribution inevitably collapses to a single point. In extreme cases this happens already after less than $10$ generations. Additionally, we find that the speed at which diversity degenerates depends on the data cycle as well as on the proportion of real and synthetic data. Fresh real data added during the data cycle slows down but can not negate the effects of a self-consuming training loop.

In summary, our main contributions are as follows:
\begin{itemize}
    \item The first comprehensive empirical study of the self-consuming training loop for LLMs trained from scratch,
    \item A novel method to unambiguously evaluate quality/correctness and diversity of LLM-generated outputs,
    \item An in-depth analysis and discussion of the effects and implications of this self-consuming training loop for LLMs.
\end{itemize}

Following this introduction, Sect.~\ref{sec:self_consuming_llms} describes the self-consuming training loop  and gives an overview of related work. In Sect.~\ref{sec:method}, we describe our experimental setting. Section~\ref{sec:results} presents the experimental results, followed by a discussion in Section~\ref{sec:discussion}. Section~\ref{sec:conclusion} concludes the paper.

\section{The Self-Consuming Training Loop}
\label{sec:self_consuming_llms}

\begin{figure}[t]
\vskip 0.2in
\begin{center}
\centerline{\includegraphics[width=\columnwidth]{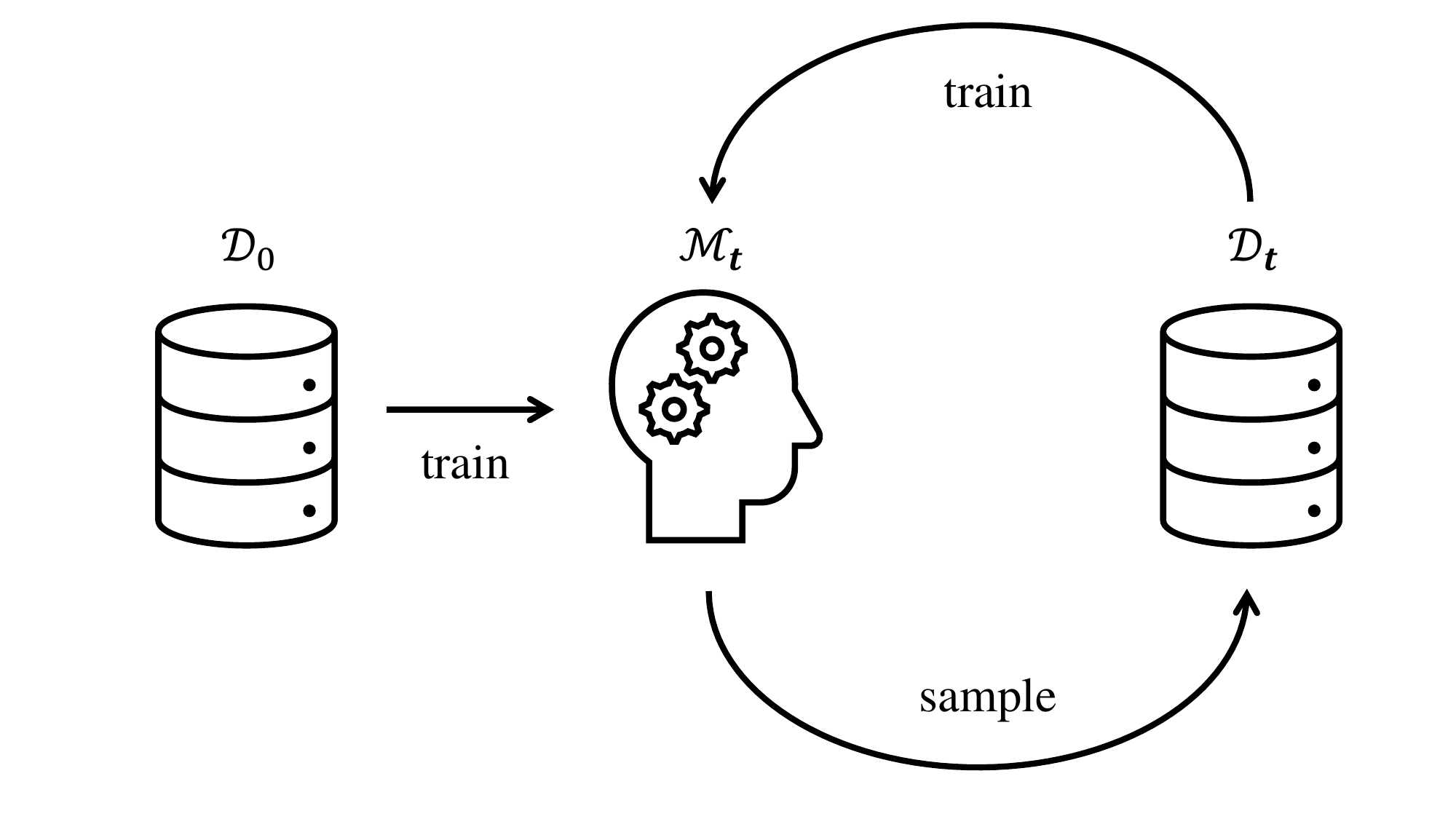}}
\caption{Self-consuming training loop: In the first generation $t = 1$ a model $\mathcal{M}_t$ is initially trained on a real dataset $\mathcal{D}_0$. From this model $\mathcal{M}_t$ a sample $\mathcal{S}_t$ is drawn to build a new dataset $\mathcal{D}_t$. 
The new dataset $\mathcal{D}_t$ is in turn used to train a new model $\mathcal{M}_{t+1}$  from scratch for the next generation $t+1$. This process is repeated iteratively until the maximum number of generations $T$ is reached.}
\label{fig:self_consuming_loop}
\end{center}
\vskip -0.2in
\end{figure}

\begin{algorithm}[tb]
   \caption{Self-Consuming Training Loop}
   \label{alg:self_consuming_loop}
\begin{algorithmic}
   \State {\bfseries Input:} number of generations $T$,
                            data cycle
    \State \textbf{Initialize:} Sample $\mathcal{D}_0$ from $P_X$
   \For{$t=1$ {\bfseries to} $t=T$}
       \State Train Model $\mathcal{M}_t$ with $\mathcal{D}_{t-1}$
       \State Sample $\mathcal{S}_t$ from $\mathcal{M}_{t}$
       \State Create $\mathcal{D}_t$ from $\mathcal{D}_{0}$ and $\mathcal{S}_{1 \dots t}$ according to 
       \State data cycle 
   \EndFor
\end{algorithmic}
\end{algorithm}

Current generations of LLMs are usually trained on large amounts of unstructured data like text and code gathered from the Internet \cite{brown2020language}. Due to their generative nature and capacity those models can in turn be used to generate new \textit{synthetic} data, often indistinguishable from the original \textit{real} data \cite{sadasivan2023can}. 
This synthetic data ends up back on the internet and thus in the next large dataset, which is used to train the next generation of LLMs.
These LLMs in turn produce new content and data, setting in motion a repetitive cycle in which new generations of models are trained each time with a higher proportion of synthetic data from previous generations.
We call this process a self-consuming training loop, depicted in Figure~\ref{fig:self_consuming_loop}.

More specifically, consider a dataset $\mathcal{D}_0$ consisting of real data points $x \in X$ sampled from the original distribution $P_X$.
A generative model $\mathcal{M}_t$ is trained on this original dataset $\mathcal{D}_0$ until the data is sufficiently fitted, producing the first generation $t=1$ of generative models.
In a self-consuming training loop,
a new set of $m$ synthetic data points $\mathcal{S}_t$ is now sampled from the previous generation of generative models $\mathcal{M}_{t}$.
The next generation of generative models $\mathcal{M}_{t+1}$ is then trained from scratch on the new dataset $\mathcal{D}_t$.
This self-consuming training loop is repeated until the maximum number of generations $t=T$ is reached.
Algorithm~\ref{alg:self_consuming_loop} presents a pseudo-code describing this self-consuming training loop.

\textbf{Related Work} Current research has analyzed the self-consuming training loop mostly in the context of image generating models.
\citet{martinez2023towards} first studied the self-consuming training loop for diffusion models and find that the self-consuming training loop leads to a collapse in diversity of the generated images.
Other work have given theoretical frameworks for this phenomenon and look at more complex data cycles in the context of image generation \cite{alemohammad2023self, shumailov2023model, bertrand2023stability}. Both \citet{alemohammad2023self} and \citet{bertrand2023stability} find that fresh real data can lead to stability within the self-consuming training loop. 

\citet{shumailov2023model} observe a degeneration of diversity for  Variational Autoencoders and Gaussian Mixture Models if some of the output is used again as input. Additionally, the authors perform experiments for iteratively fine-tuning LLMs in a self-consuming way and observe a degradation in quality. They find that degradation is less strong than in the imagine generation context and that keeping some original data helps mitigating this phenomenon. Other contemporary work also analyzed repeated fine-tuning of LLMs in a self-consuming way and observed a decrease in lexical diversity \cite{guo2023curious}.

However, new generations of LLMs are usually trained from scratch with web scraped datasets while fine-tuning is mainly done with curated datasets. So the analysis of the self-consuming training loop when training from scratch is of pressing concern. To the best of our knowledge, we are the first to study the behavior of LLMs trained in a self-consuming loop from scratch. Furthermore, self-consuming loops in LLMs have not yet been analyzed with a method that allows to unambiguously evaluate the quality and diversity of generated model output.

\begin{figure*}[t]
    \centering
        \subfigure[Full Synthetic Data Cycle]{\includegraphics[width=0.66\columnwidth]{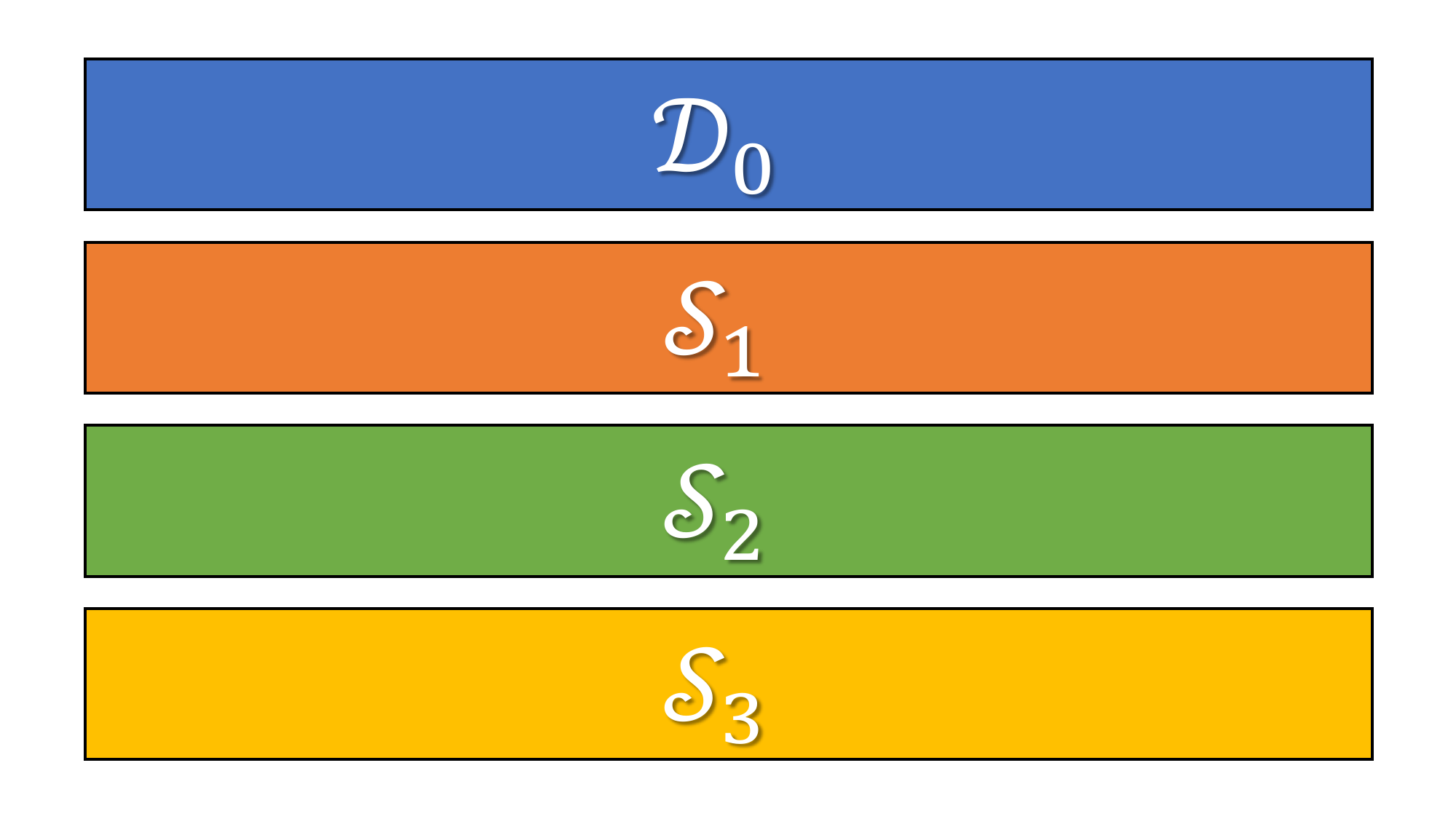}}
        \subfigure[Balanced Data Cycle]{\includegraphics[width=0.66\columnwidth]{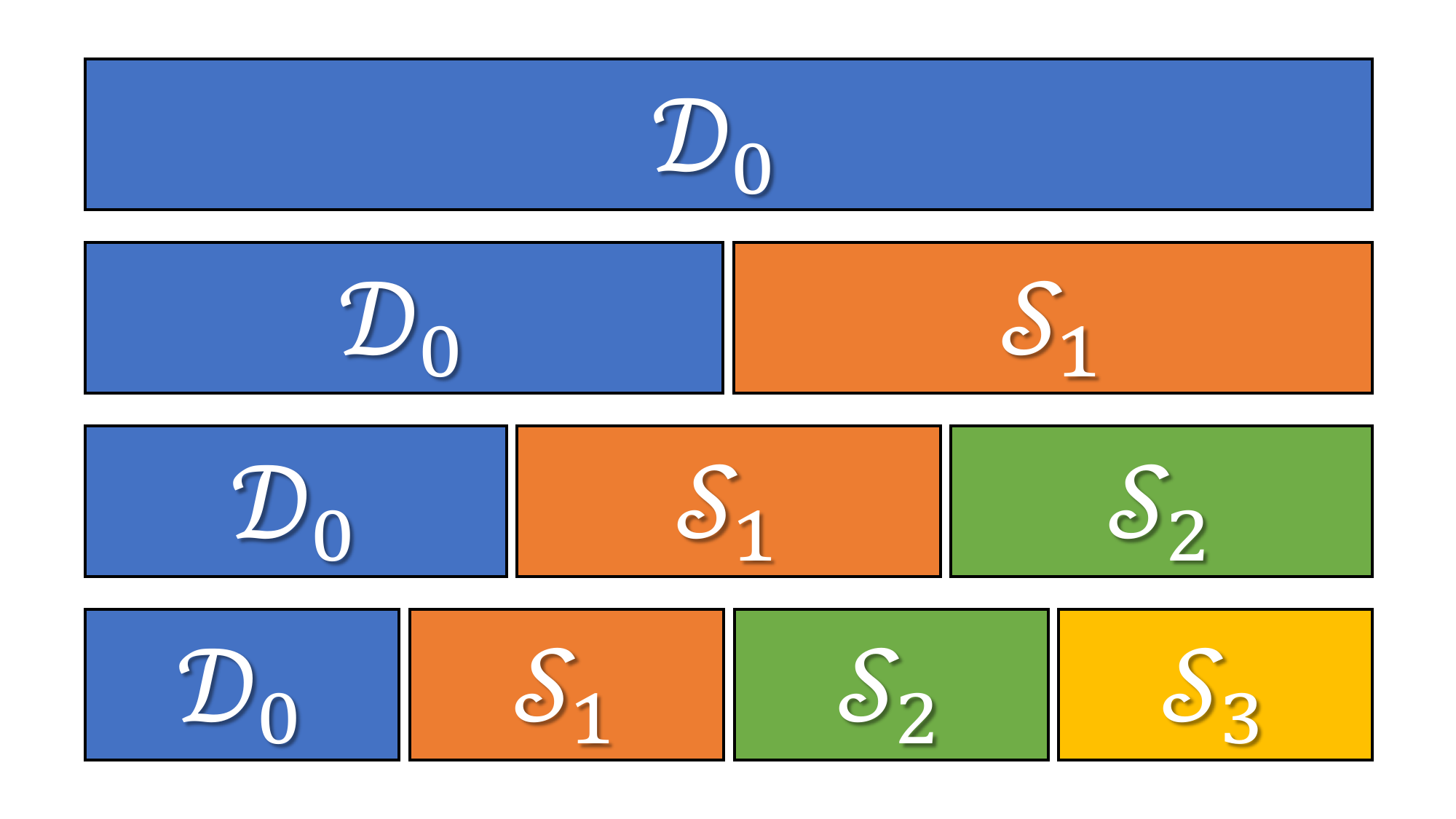}}
        \subfigure[Incremental Data Cycle]{\includegraphics[width=0.66\columnwidth]{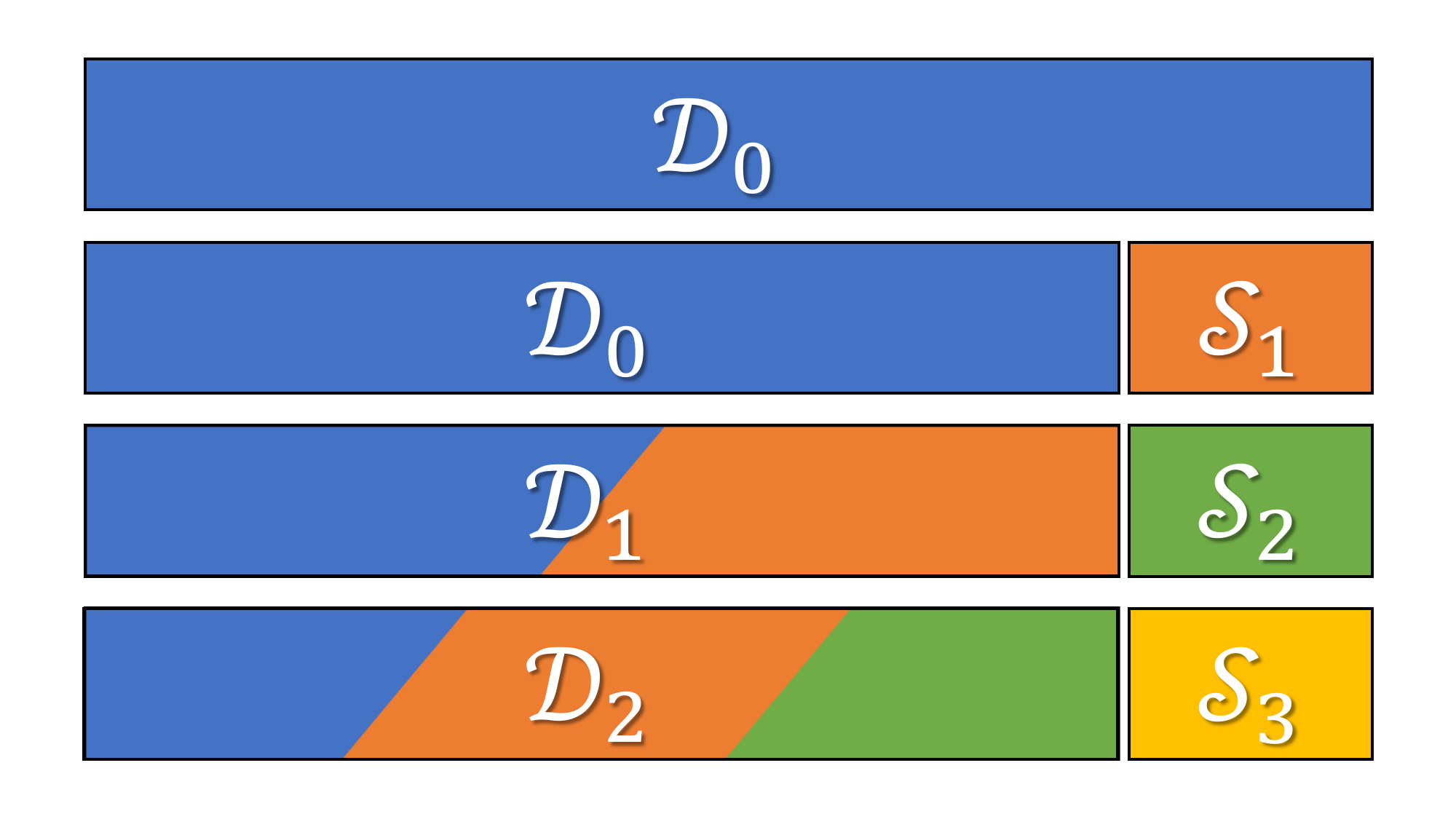}}
        \subfigure[Expanding Data Cycle]{\includegraphics[width=1.94\columnwidth]{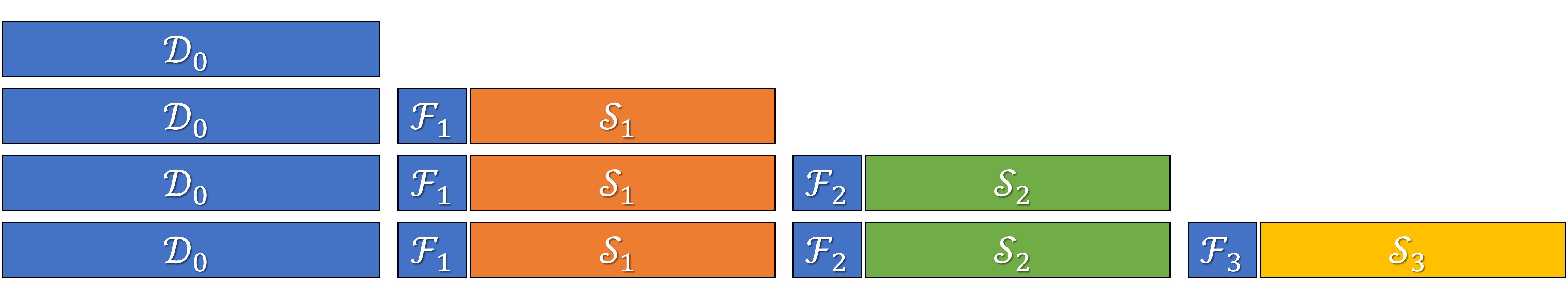}}
    \caption{Each sub figure represents a different data cycle exemplary for four generations. The first row in each sub figure is the original dataset $\mathcal{D}_0$. The second, third and fourth row depict the dataset $\mathcal{D}_1$, $\mathcal{D}_2$, and $\mathcal{D}_3$ respectively.} 
    \label{fig:data_cycles}
\end{figure*}

\section{Experimental Setup}
\label{sec:method}

We present our logic expression dataset, which is the foundation for the verification of the language model's output. Furthermore, we explain the data cycles we analyze in our experiments and describe the used model architecture in detail. 

\subsection{Verification with a Logic Expression Dataset}
\label{subsec:logic_expressions}

LLMs usually generate natural language texts and their performance is typically measured by using similarity metrics like the BLEU score \cite{papineni2002bleu}, ROUGE score \cite{lin2004rouge} and BERT score \cite{zhang2019bertscore} as well as perplexity \cite{jurafsky2009speech}. However, those metrics rely on measuring similarity of outputs with expected reference data, which can only serve as a proxy for quality of a language model \cite{callison2006re, gehrmann2023repairing}. 
Consequently, we propose using a dataset consisting of logic expressions. Those expressions can be represented as a sequence making them a good fit for language modeling. 
In contrast to  natural language text, we can easily and systematically evaluate the quality of logic expressions by verifying their correctness. 
An expression is defined to be syntactically correct, if  it can be parsed without an error. The  semantic correctness can be evaluated if an expression is either \texttt{True} or \texttt{False}. If the original dataset only consists of \texttt{True} expressions, then a high-quality trained model should also only generate \texttt{True} expressions.

We build a logic expression as a tree in a recursive way specified in the function \texttt{GenLogicExpr($d$)}, where $d$ is the desired depth of a logic expression tree. If the desired tree depth is reached when calling the function, a random Boolean is returned (either \texttt{True} or \texttt{False}). Otherwise, a logic operator (\texttt{not,and,or}) is selected and the function is called again recursively with $d-1$.
To build the entire original dataset we use this recursive function to sample new random logic expression trees with a random depth between $d_{min}$ and $d_{max}$ until we have a dataset with $m$ unique expressions that evaluate in a \texttt{True} Boolean expression. The complexity of the dataset can be easily controlled by adjusting $d_{max}$.
For our experiments we chose a initial dataset size of $m = 10,000$ with expressions of minimum depth $d_{min} = 1$ and maximum depth $d_{max} = 5$. 
Algorithm~\ref{alg:generate_logic_expr} in Appendix~\ref{appendix:gen_logic_expr} describes the generation of our logic expression dataset in pseudo-code.

The resulting logic expression trees can then be saved as strings, e.g., \texttt{not ( True and False~)}. 
This allows us to evaluate them in Python using the \texttt{eval()} function to test whether  they are correct or trigger an error and whether they evaluate to a \texttt{True} or \texttt{False} Boolean.
Additionally, we can encode those strings to a sequence of tokens (\texttt{True, False, not, and, or, (, ), <eos>}, where \texttt{<eos>} is a stop token and indicates the end of an expression) and use these sequences to train a language model.

\subsection{Measuring the Diversity of the Model's Output}
\label{subsec:diversity_measure}

A generative model does not only need to generate correct outputs but also a diverse set of outputs.
To measure the diversity of a sample from a LLM we use the Levenshtein diversity \cite{wittenberg2023denoising} in our experiments. This metric calculates the pairwise Levenshtein distance between each expression in the sample normalized by the number of tokens of the longer expression of each pairwise comparison and averaged over the number of pairwise comparisons. If the average pairwise normalized Levenshtein distance is close to zero the diversity within a sample is low. A value closer to one suggests a stronger diversity within a sample. The Levenshtein distance itself is a metric for measuring the distance between two strings and is defined as the number of edits (deletion, addition, substitution) required to change one string to another \cite{levenshtein1966binary}. Normally, this edit distance is calculated on a character basis, however, we calculate it on the tokens from our encoding of the logic expressions to accommodate for the different length each token has in character representation.

\subsection{Data Cycles}
\label{subsec:data_cycle}

We define a \textit{data cycle} as the way a dataset $\mathcal{D}_t$ is constructed in generation $t$ from the original data $\mathcal{D}_0$, potential fresh data $\mathcal{F}_t$ from the original distribution, and the generated data from the current and previous generations $\mathcal{S}_{1 \dots t}$. This way of constructing a new dataset may vary and different data cycles are possible.
In the image generation domain, previous work suggests that the self-consuming training loop is influenced by the data cycle being used \cite{alemohammad2023self}. Therefore, inspired by their work we conduct our experiments with the following four different data cycles (also depicted in Fig.~\ref{fig:data_cycles}):

\textbf{Full Synthetic Data Cycle:} 
In the most extreme case of a self-consuming training loop a new model $\mathcal{M}_t$ is only trained on the generated data from the last generation so that $\mathcal{D}_{t-1} = \mathcal{S}_{t-1}$. We call this a \textit{full synthetic} data cycle. 
Normally, new datasets would still contain the original data when they are collected in practice. 
However, we can use this data cycle to study the most extreme changes in behavior of models from generation to generation.

\textbf{Balanced Data Cycle:}
We refer to the second data cycle as the \textit{balanced} data cycle.
In this data cycle, we construct the new dataset $\mathcal{D}_t$ from equal parts of all the previous samples $\mathcal{S}_{1 \dots t}$ and the original dataset $\mathcal{D}_0$ so that every previous generation contributes $m * \frac{1}{t+1}$ logic expressions to the new dataset $\mathcal{D}_t$ of size $m$.

\newpage
\textbf{Incremental Data Cycle:}
The third data cycle we study is called the \textit{incremental} data cycle. In this data cycle, the new dataset is created by a portion $(1-\lambda)$ of the last dataset $\mathcal{D}_{t-1}$ and a portion $\lambda$ of new sampled data $\mathcal{S}_t$ so that $\mathcal{D}_t = (1 - \lambda) * \mathcal{D}_{t-1} + \lambda * \mathcal{S}_t$ while holding the size $m$ of the dataset constant each generation. A $\lambda = 1$ would result in a full synthetic data cycle. We chose $\lambda = 0.1$ for our initial experiments and later used different values of $\lambda$ to study the influence of this parameter on diversity.

\textbf{Expanding Data Cycle:}
The first three data cycles create a dataset of equal size in each generation. In practice, however, the dataset would grow each generation by adding new generated data to the already existing dataset. Additionally, fresh data samples $\mathcal{F}_t$ from the original distribution (e.g. human generated) would also be added at each generation. Consequently, the last data cycle we study is an \textit{expanding} data cycle. At each generation the new dataset is created so that $\mathcal{D}_t = \mathcal{D}_{t-1} + (1-\lambda) * \mathcal{F}_t + \lambda * \mathcal{S}_t $, where $\lambda$ is the portion of generated data we add each generation and $(1-\lambda)$ is the portion of fresh real data added at each generation. For this data cycle, we chose $\lambda = 0.9$ for our initial experiments and later used different values of $\lambda$ to study the influence of fresh data in an expanding data cycle on diversity.

\begin{figure*}[ht]
    \centering
        \subfigure[Full Synthetic Data Cycle]{\includegraphics[width=0.95\columnwidth]{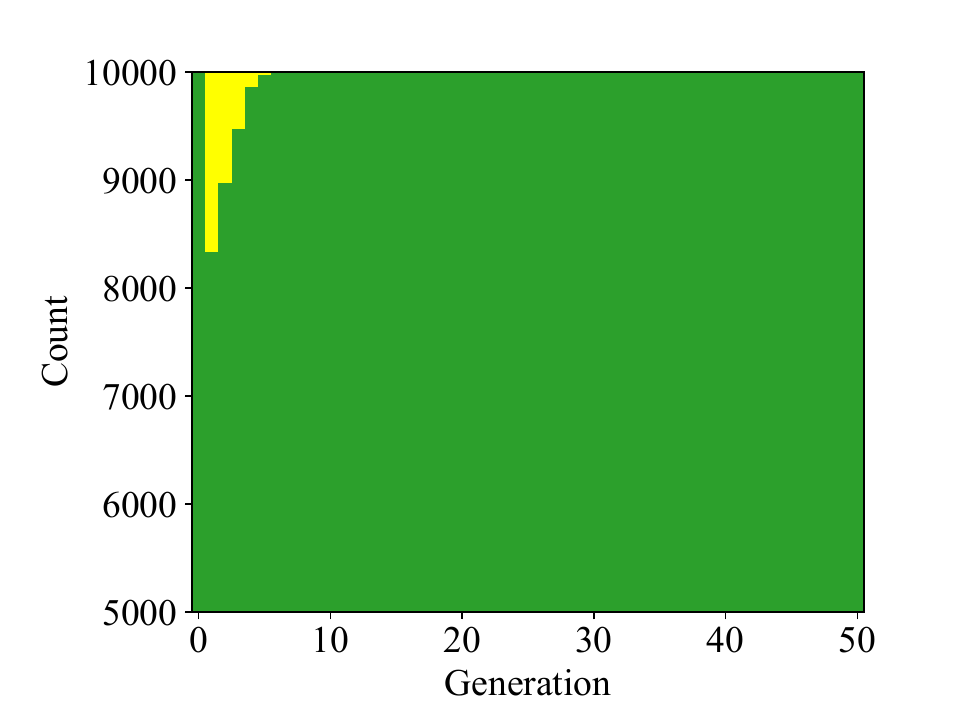}}
        \subfigure[Balanced Data Cycle]{\includegraphics[width=0.95\columnwidth]{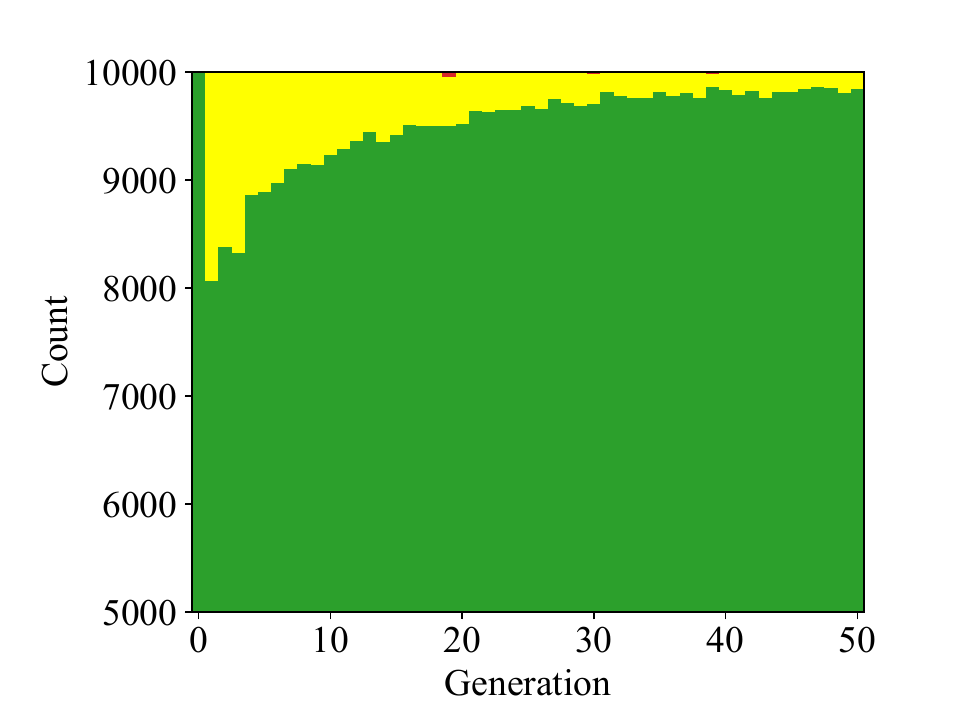}}
        \subfigure[Incremental Data Cycle]{\includegraphics[width=0.95\columnwidth]{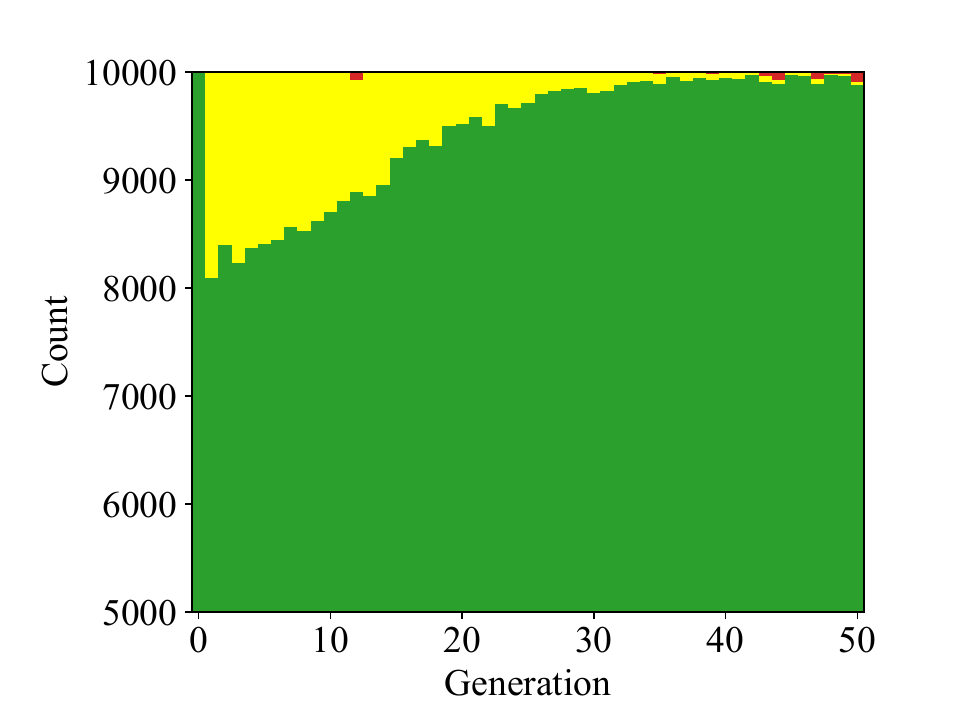}}
        \subfigure[Expanding Data Cycle]{\includegraphics[width=0.95\columnwidth]{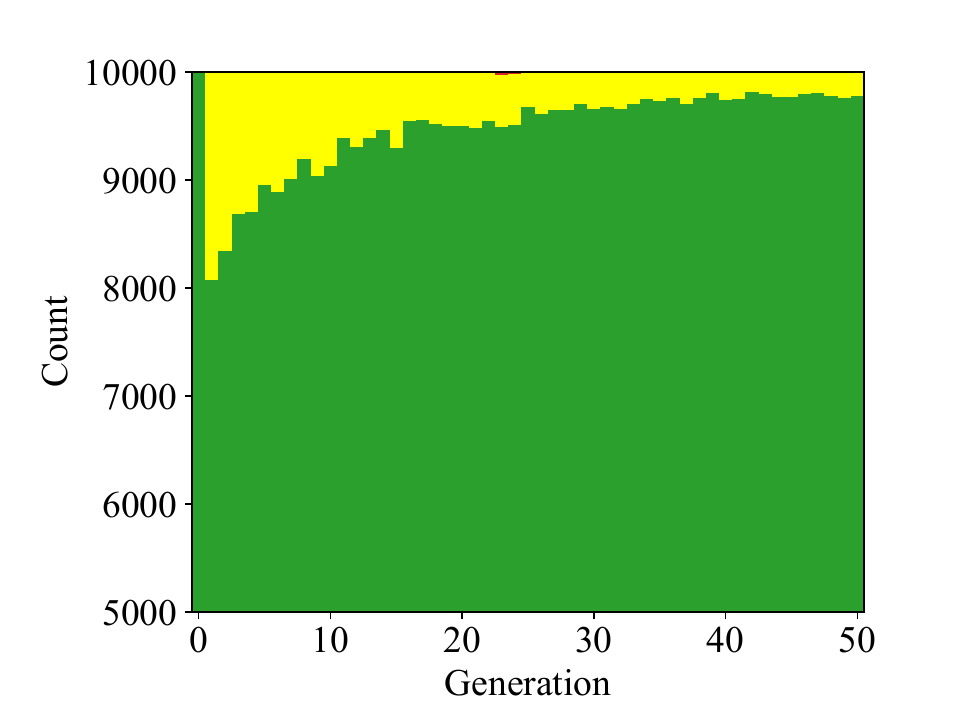}}
        \subfigure{\includegraphics[width=1\columnwidth]{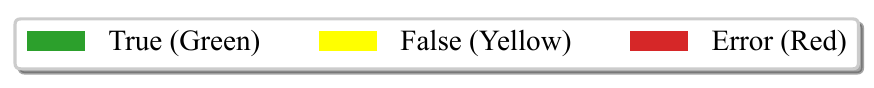}} 
    \caption{The composition of each sample $\mathcal{S}_t$ from model $\mathcal{M}_t$ at generation $t$ (the first bar displays the composition of $\mathcal{D}_0$) with regards to the number of syntactically and semantically correct expressions. Green indicates syntactically correct expressions that evaluate to \texttt{True}, yellow indicates syntactically correct expressions that evaluate to \texttt{False}, and red indicates syntactically incorrect expressions that result in an error when being parsed. Each subplot displays the results for a different data cycle.}
    \label{fig:quality}
\end{figure*}

\subsection{Model Architecture and Training}
\label{subsec:model}

We employ a GPT-style LLM using the open-source implementation \textit{nanoGPT}\footnote{\href{https://github.com/karpathy/nanoGPT}{https://github.com/karpathy/nanoGPT}} (MIT license) in our experiments. The model accepts a context of up to 256 tokens and consists of 6 attention layers with 6 attention heads each and an embedding dimensionality of 384, resulting in roughly $10.6$ million parameters.

During training we use a batch size of 64 and a dropout rate of $0.2$, training for 5000 iterations minimizing cross entropy loss, starting with a learning rate of $10^{-3}$ decaying to $10^{-4}$, to achieve a sufficient fitting of the training data. We split the dataset in 90\% training and 10\% validation data. During training we calculate the validation error every 250 iterations and use the model parameters with the lowest validation error as our final model $M_t$ for each generation $t$ in the self-consuming training loop.

We use the trained model at each generation to sample $m = 10,000$ logic expressions. During sampling, we auto-regressively generate new tokens with temperature $0.8$ and feed them back into the model until a stop token \texttt{<eos>} is sampled up to a maximum of 200 tokens per expression. We run the self-consuming training loop for $T=50$ generations in each experiment.

All experiments are performed on a workstation using a NVIDIA Turing architecture graphics card. 

\section{Results}
\label{sec:results}

We present our experimental results. First, we describe our results on the performance in terms of correctness of models trained in a self-consuming training loop. Afterwards, we present our results in terms of diversity for those models.

\subsection{Correctness of Generated Content}
\label{subsec:quality}

We first study the correctness of the expressions within a generated sample $\mathcal{S}_t$ from a model $\mathcal{M}_t$ at each generation $t$. As described in Sec.~\ref{subsec:logic_expressions}, we consider an expression to be syntactically correct if it can be parsed without error. Since the original dataset $\mathcal{D}_0$ only consists of \texttt{True} expressions, we consider a generated expression to be semantically correct if it also evaluates to \texttt{True}. A semantically correct expression is also syntactically correct.

Figure~\ref{fig:quality} displays the composition of samples $\mathcal{S}_t$ generated during a self-consuming training loop. Each subplot presents the results for a different data cycle: a) full synthetic, b) balanced, c) incremental, and d) expanding. Every bar in a subplot displays the composition of $\mathcal{S}_t$ generated from model $\mathcal{M}_t$ at generation $t$, except for the first bar in each subplot which displays $\mathcal{D}_0$. The green portion of a bar indicates the number of syntactically correct expressions that evaluate to \texttt{True} in that sample. The yellow part of a bar represents the number of syntactically correct expressions that evaluate to \texttt{False}, and the red part of a bar displays the number of syntactically incorrect expressions that result in an error when being parsed.

Overall, we see that only very few expressions are syntactically incorrect. This indicates that the models are sufficiently trained and can correctly learn the syntactic rules of a logic expression.
We see a drop of around 20\% in the number of semantically correct expressions from the original data $\mathcal{D}_0$ to the first sample $\mathcal{S}_1$ in every data cycle. 
Interestingly, the number of semantically correct expressions increases afterwards over the course of the self-consuming training loop.
The speed at which this number increases depends on the data cycle. We can see the fastest increase for the full synthetic data cycle in which the whole sample consists of \texttt{True} expressions by generation $t=6$. The incremental data cycle takes consistently longer but also nearly reaches this point by generation $t=36$. While the balanced data cycle has an initially steeper increase in semantically correct expressions, it does not reach the point of a completely semantically correct sample by the end of 50 generations but comes very close to it. Lastly, the expanding data cycle shows this trend as well. 
Overall, we see that the self-consuming training loop seems to help with generating more semantically correct expressions in our experiments and the rate at which this happens differs by data cycle.

\subsection{Diversity of Generated Content}
\label{subsec:diversity}

\begin{figure}[t]
\vskip 0.2in
\begin{center}
\centerline{\includegraphics[width=\columnwidth]{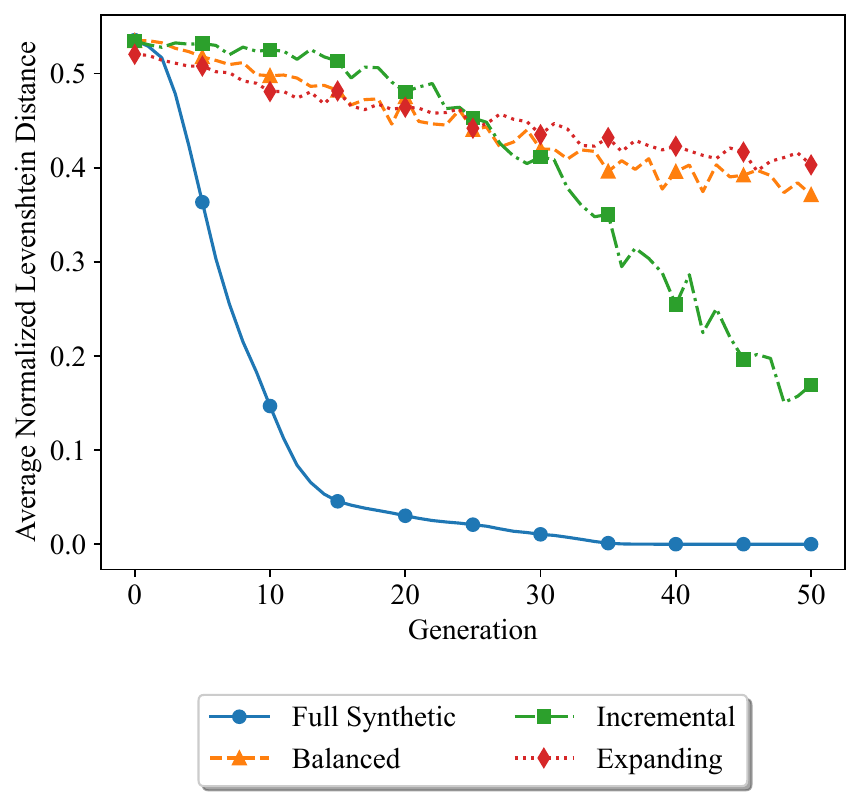}}
\caption{Average pairwise normalized Levenshtein distance over generations of a self-consuming training loop for different data cycles. Line markers added every 5 generations for better display. }
\label{fig:diversity_per_data_cycle}
\end{center}
\vskip -0.2in
\end{figure}

\begin{figure}[t]
\vskip 0.2in
\begin{center}
\centerline{\includegraphics[width=\columnwidth]{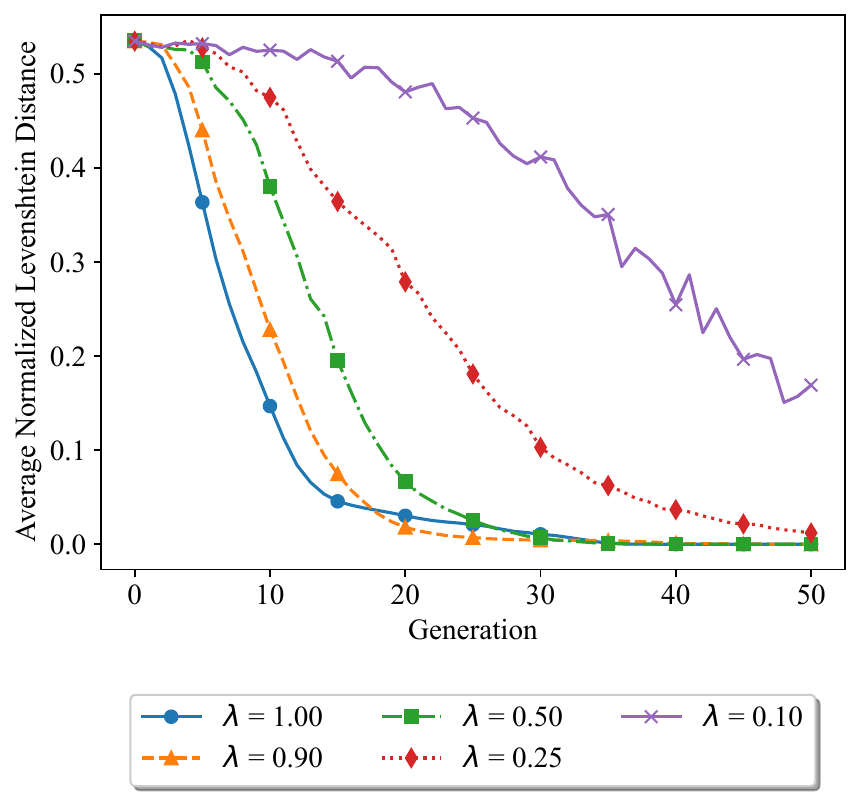}}
\caption{Average pairwise normalized Levenshtein distance over the course of a self-consuming training loop for the incremental data cycle with different portions $\lambda$ of new sampled data. Line markers added every 5 generations for better display.}
\label{fig:diversity_per_percentage_generated_data}
\end{center}
\vskip -0.2in
\end{figure}

\begin{figure}[t]
\vskip 0.2in
\begin{center}
\centerline{\includegraphics[width=\columnwidth]{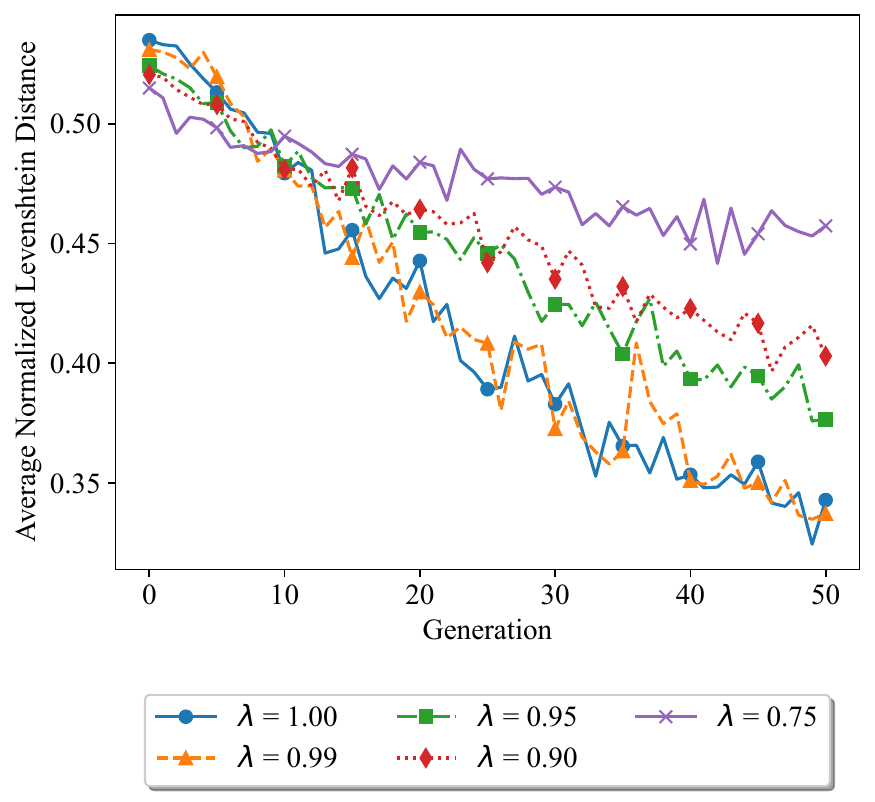}}
\caption{Average pairwise normalized Levenshtein distance over the course of a self-consuming training loop for the expanding data cycle for different portions $\lambda$ of generated data and $(1-\lambda)$ of fresh data. Line markers added every 5 generations for better display.}
\label{fig:diversity_per_percentage_generated_data_with_fresh_data}
\end{center}
\vskip -0.2in
\end{figure}

While we can see an increase in correctness over generations in our experiments, contemporary work in the image generation domain suggests, that this comes with a loss of diversity \cite{martinez2023towards, alemohammad2023self, bertrand2023stability}. Therefore, in this section we study the diversity within a sample $\mathcal{S}_t$ of model $\mathcal{M}_t$ for each generation $t$.

Figure~\ref{fig:diversity_per_data_cycle} displays the average pairwise normalized Levenshtein distance over generations of a self-consuming training loop for different data cycles. 
For each data cycle, we observe a decrease in diversity over the course of generations, with the degree of decrease varying depending on the data cycle.
For the full synthetic data cycle we see a steep decline in diversity with a collapse into a single point by generation $t=39$.
The incremental data cycle is initially stable, but decreases in diversity from the 10th generation onwards, with a decrease of 68\% in diversity by the end of the 50 generations.
The balanced and expanding data cycle also decrease in diversity, however, this decrease is way slower, with a 30\% decrease in diversity for the balanced data cycle and a 22\% decrease for the expanding data cycle.  
While not all data cycles fully collapse in diversity by generation 50, ultimately, we expect all of them to eventually reach zero diversity if the self-consuming training loop is run for enough generations.

To get further insight into the decline in diversity, we also inspect the number of unique expressions sampled per generation for the full synthetic data cycle. We find that the number of unique expressions is stable at first while the diversity already declines and then rapidly decreases within 5 generations to very few unique expressions
(see Appendix~\ref{appendix:uniques}).

To better understand the influence of the amount of generated data in each generation on the diversity, we also study the incremental data cycle in more detail with different portions $\lambda$ of new sampled data.
Figure~\ref{fig:diversity_per_percentage_generated_data} displays the average pairwise normalized Levenshtein distance over generations in a self-consuming training loop for the incremental data cycle with different values of $\lambda$.
We observe, that the loss in diversity is stronger for a higher share of  generated data added in each generation of the self-consuming training loop. Even with as little as $\lambda = 0.25$, we reach the complete loss of diversity within $50$ generations. Only for our experimental run with $\lambda = 0.10$ the diversity does not drop to zero, but to a value around $0.17$ after $50$ generations. However, we see that the speed at which diversity decreases does not scale linear with the amount of generated data. Instead we observe a more exponential increase in speed where larger amounts of generated data lead to an even quicker decay in diversity. 

Lastly, we study the expanding data cycle in more detail with regard to the proportion between generated data and fresh data added to the training data at each generation. 
Figure~\ref{fig:diversity_per_percentage_generated_data_with_fresh_data} displays the average pairwise normalized Levenshtein distance over generations in a self-consuming training loop for the expanding data cycle with different portions $\lambda$ of generated data and $(1-\lambda)$ fresh real data.
We observe, that while each configuration declines in diversity, the rate of this decline depends on the proportion between generated and fresh data. The more fresh data is added at each generation, the slower diversity decreases. With no fresh data or only 1\% of fresh data the diversity decreases by approximately  36\%. If the portion of fresh data is 25\% we only observe a decrease in diversity of approximately 11\%. This indicates that fresh data cannot stop the self-consuming training loop but enough fresh data can slow down the rate of the decline in diversity.

\section{Discussion}
\label{sec:discussion}

In our experiments we found that iteratively re-training LLMs from scratch might help with correctness of model output. This is sometimes already done in practice where the output of LLMs is used to improve their performance \cite{huang2022large, wang2022self} or generate new training data \cite{yu2023large}.
However, this trend is concerning as we also found that this self-consuming training loop eventually leads to a drastic loss of diversity of a model's output.
Our results confirm the findings on the self-consuming training loop in the field of image generation \cite{martinez2023towards,alemohammad2023self,shumailov2023model,bertrand2023stability}, further highlighting the importance of this issue for generative models and LLMs in particular.
These LLMs have an even greater impact on society than image generation models, leading to more LLM-generated data being produced and potentially having a greater impact on society as a whole, as many people already rely on good model performance.

Consequently, as we expect that a self-consuming training loop will harm model performance in the future, researchers and practitioners should carefully choose their data when training their models and test those models sufficiently for diversity. Additionally, the machine learning community at large needs to find ways to deal with this problem as generated text data is already present in new internet scraped datasets. 
While our results indicate that fresh real data can slow down the self-consuming training loop, other studies suggest that this is not an option as we will run out of real data eventually and have to rely on generated data for future models \cite{villalobos2022will}. 
Additionally, LLM-generated data has already found its way onto the internet, even infecting academic publications \cite{geng2024chatgpt,liang2024mapping}, and LLM-generated data may soon overtake real data on the internet.

Furthermore, this LLM-generated data is hard to differentiate from real data \cite{kreps2022all, sadasivan2023can, huschens2023you}, making it difficult to curate datasets scraped from the web. This makes the self-consuming training loop inevitable in the future. Therefore, we advise to further study ways to maintain diversity of generative model outputs like quality diversity methods \cite{pugh2015confronting, fontaine2021differentiable}.

\section{Conclusion}
\label{sec:conclusion}

In this paper we studied the behavior of LLMs trained in a self-consuming training loop and found that iteratively training LLMs from scratch with self generated data can initially help with correctness of model outputs. However, this comes at a price, as the diversity of generated data eventually degenerates and collapses into a single point. The rate of this decline in diversity depends on the data cycle creating the training data at each generation. Furthermore, we found that fresh data can slow down this process and preserve diversity of a model output for longer. This should encourage researches to carefully look at their datasets and models to mitigate such a harmful self-consuming training loop in the future.

In future work, we will study how fine-tuning can influence the self-consuming training loop for LLMs. Additionally, we plan to further investigate how the effects of a self-consuming training loop can be slowed down or completely negated.

\newpage
\section{Limitations}
\label{sec:limitations}

One major cause for limitations of our work are limited computational resources. First, we ran our experiments with a smaller GPT-style model than for example GPT-4. Even though we expect that larger models with billions of parameters still suffer from the same loss of diversity if trained in a self-consuming training loop, we note that the behavior may vary for those larger models. However, a study for models of those size is not feasible with current compute options nor is it responsible in terms of the expected CO$_2$ emissions \cite{strubell2019energy}.

Second, another limitation imposed by our computational resources is that the main results of this work consist of only one run per experimental configuration. With more runs per experiment the results would be more robust, but conducting more runs for our experiments was simply not feasible with our available computing resources. As all our experiments point towards the same direction, however, we do not believe that this limitation is crucial.
Nevertheless, to further mitigate this limitation we performed additional experiments for the full synthetic data cycle with fewer generations over multiple runs and with different initializations in Appendix~\ref{appendix:init_impact}. These results show that the trend of a decline in diversity still holds true over multiple runs.

Another limitation is that our work focuses on repeatedly training foundation models from scratch in a self-consuming way. However, in practice those models are normally fine-tuned after the initial training. It is unclear from our work if this fine-tuning process can slow down or negate the effects of a self-consuming training loop when performed with fresh real data. However, we plan to investigate this in future work.

Lastly, one may argue that real text has a way larger vocabulary and more complex structures. Therefore, our results might be slightly different when performed on a real text corpus. However, measuring both quality and diversity of generated text is not trivial and can only achieved by using a proxy \cite{callison2006re, gehrmann2023repairing}. Consequently, we intentionally performed our study on a logic expression dataset to ensure a controlled experimental setting in which we can measure both quality and diversity of a trained model.

\bibliography{custom}

\newpage
\appendix

\section{Generation of the Logic Expression Dataset}
\label{appendix:gen_logic_expr}

Algorithm~\ref{alg:generate_logic_expr} describes the generation of our logic expression dataset in detail.

\begin{algorithm}[!ht]
\caption{Generate Logic Expression Dataset}
\label{alg:generate_logic_expr}
\begin{algorithmic}
\State {\bfseries Input:} number of unique expressions $m$, 
                            minimum tree depth $d_{min}$,
                            maximum tree depth $d_{max}$

\State \textbf{Initialize:} $\mathcal{D}_0 = \{\}$ \\

\State \textbf{function} {$\textrm{GenLogicExpr}(d):$}

    \If{$d=0$}
        \State \textbf{return}  $\textrm{RandomBool}()$
    \Else
        \State $op \gets \textrm{RandomLogicOperator}()$
        \If{$op = \textrm{'not'} $}
            \State \textbf{return} $op + \textrm{GenLogicExpr}(d-1)$
        \Else
            \State ${leftExpr} \gets \textrm{GenLogicExpr}(d-1)$
            \State $rightExpr \gets \textrm{GenLogicExpr}(d-1)$
            \State \textbf{return} $leftExpr + op + rightExpr$
        \EndIf
    \EndIf
\State \textbf{end function} \\
\While{$|\mathcal{D}_0| < m$}
    \State $i \gets \textrm{RandomIntegerInRange}(d_{min}, d_{max})$
    \State $expr \gets \textrm{GenLogicExpr}(d=i)$
    \If{$\textrm{EvalBoolExpr}(expr) = \textrm{True}$}
        \State $\mathcal{D}_0 \gets \mathcal{D}_0 \cup \{expr\}$
    \EndIf
\EndWhile

\end{algorithmic}
\end{algorithm}

\section{Unique Expressions Generated}
\label{appendix:uniques}

\begin{figure}[!h]
\vskip 0.2in
\begin{center}
\centerline{\includegraphics[width=\columnwidth]{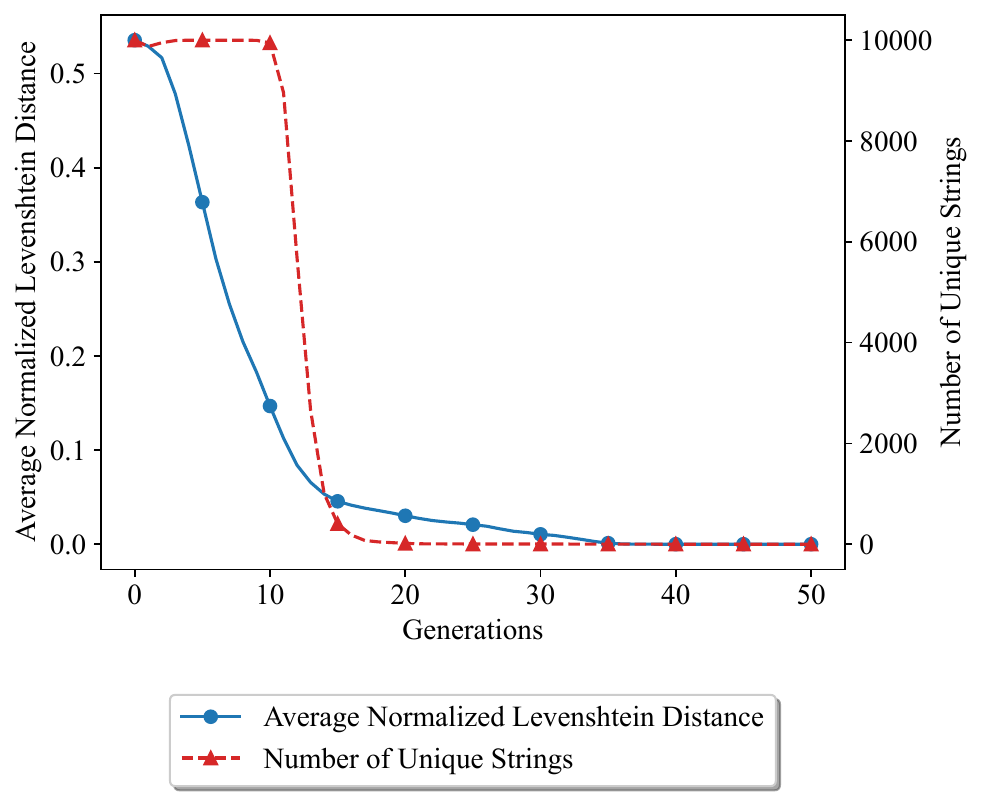}}
\caption{Average pairwise normalized Levenshtein distance (left y-axis) and number of unique expressions in a sample (right y-axis) over generations of a self-consuming training loop for the full synthetic data cycle. Line markers added every 5 generations for better display. }
\label{fig:number_uniques_full_synth}
\end{center}
\vskip -0.2in
\end{figure}

To better understand the decline of diversity, we inspect the number of unique expressions sampled at each generation. We focus on the full synthetic data cycle, as this data cycle fully collapses to a single point within 50 generations.

Figure~\ref{fig:number_uniques_full_synth} displays the number of unique expressions (right y-axis) generated at each generation of a self-consuming training loop in comparison to the diversity measured as the average pairwise normalized Levenshtein distance (left y-axis).
We observe that in the beginning the number of unique expressions stays stable while the diversity is already in a steep decline. Only by generation $t=10$ the number of unique individuals starts to decline rapidly and by generation $t=15$ only around 400 unique expressions are sampled while the decline in diversity slows down. In generation $t=21$ the number of unique individuals is below 10 and in generation $t=39$ only a single individual is sampled at each generation. Consequently, the diversity is collapsed to zero. 

This shows that it is not enough to only track unique outputs of a model, but that diversity needs to be tracked as well. By the time a decrease in unique results is noticeable, the diversity already decreased significantly.

\section{Impact of Initialization}
\label{appendix:init_impact}

\begin{figure*}[!ht]
    \centering
        \subfigure[True Initialization]{\includegraphics[width=0.66\columnwidth]{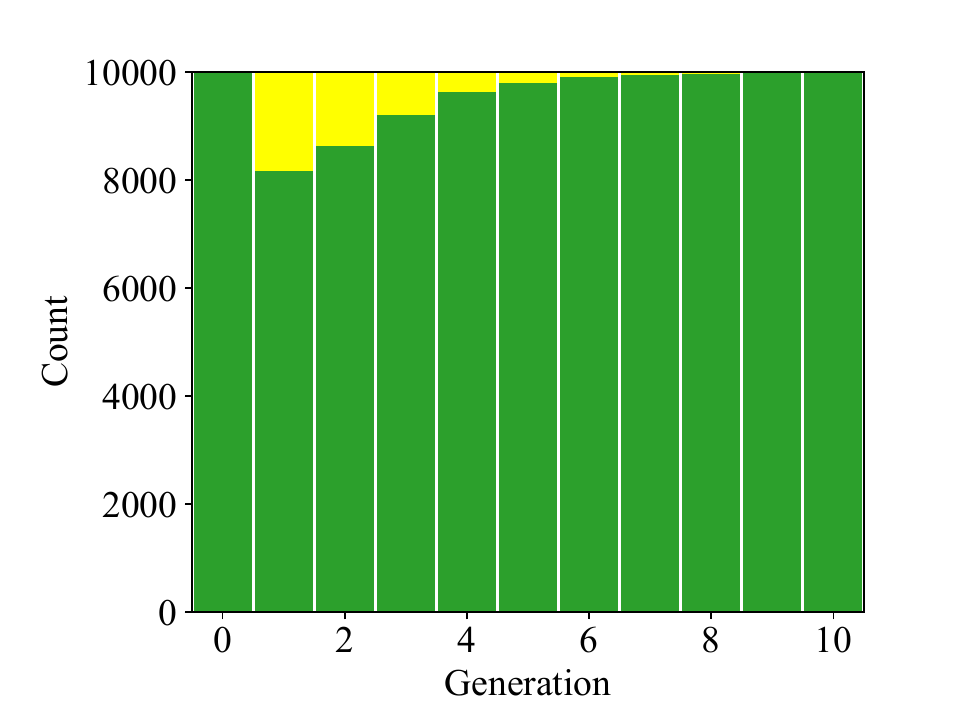}}
        \subfigure[False Initialization]{\includegraphics[width=0.66\columnwidth]{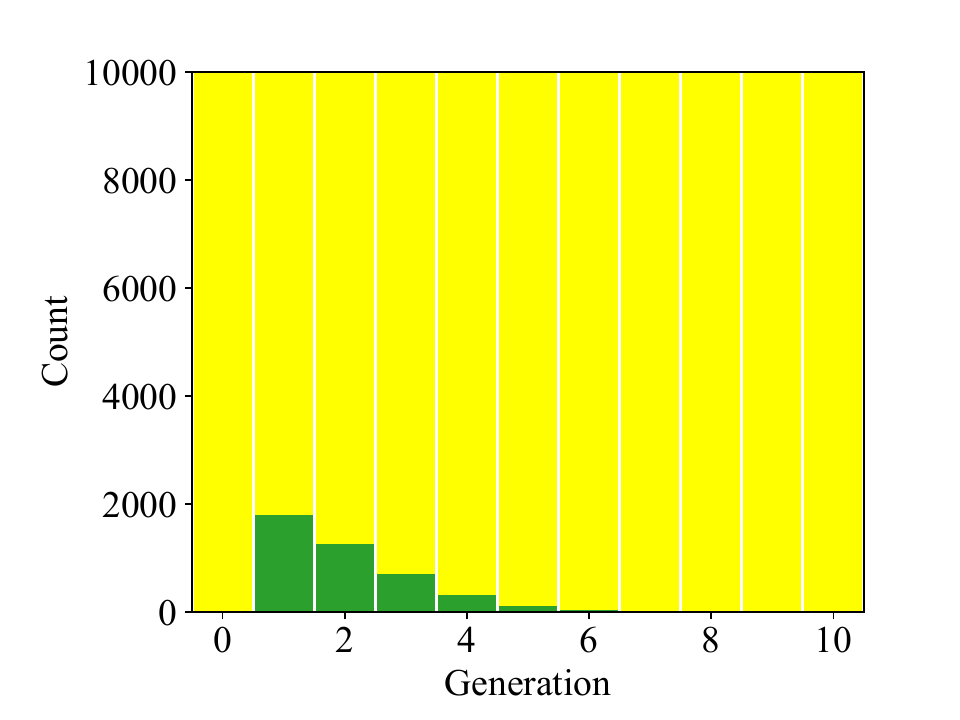}}
        \subfigure[Mixed Initialization]{\includegraphics[width=0.66\columnwidth]{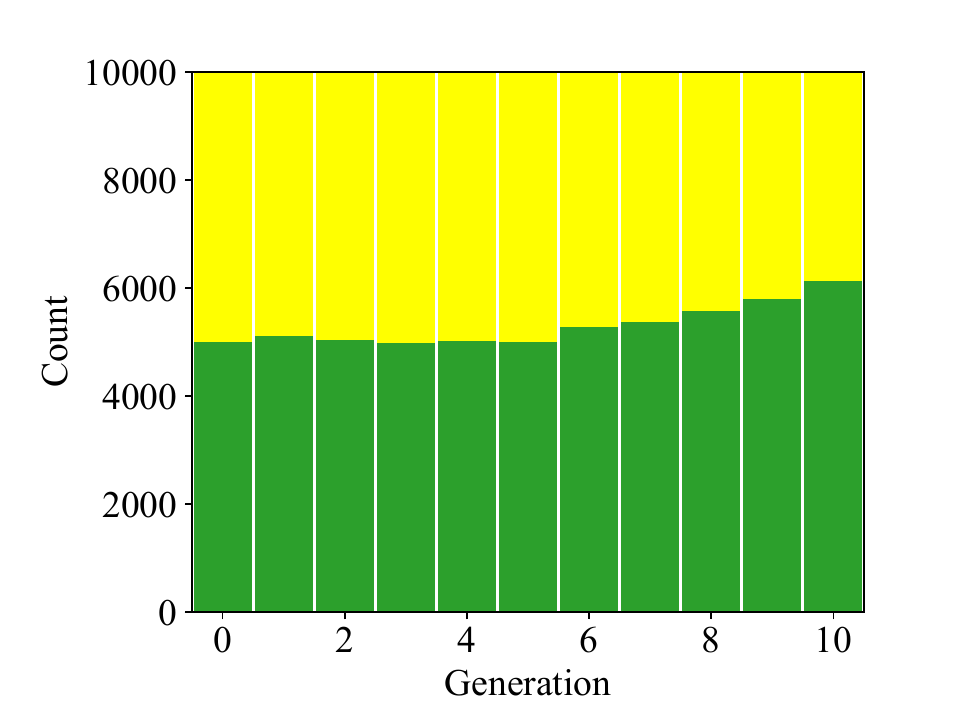}}
        \subfigure{\includegraphics[width=1\columnwidth]{figures/quality_barplot_legend.pdf}} 
    \caption{The composition of each sample $\mathcal{S}_t$ from model $\mathcal{M}_t$ at generation $t$ (the first bar displays the composition of $\mathcal{D}_0$) with regards to the number of syntactically and semantically correct expressions averaged over 5 runs. Green indicates syntactically correct expressions that evaluate to \texttt{True}, yellow indicates syntactically correct expressions that evaluate to \texttt{False}, and red indicates syntactically incorrect expressions that result in an error when being parsed. Each subplot displays the results for the full synthetic data cycle with different initializations (\texttt{True}, \texttt{False}, Mixed).}
    \label{fig:quality_per_init}
\end{figure*}

\begin{figure}[!h]
\vskip 0.2in
\begin{center}
\centerline{\includegraphics[width=\columnwidth]{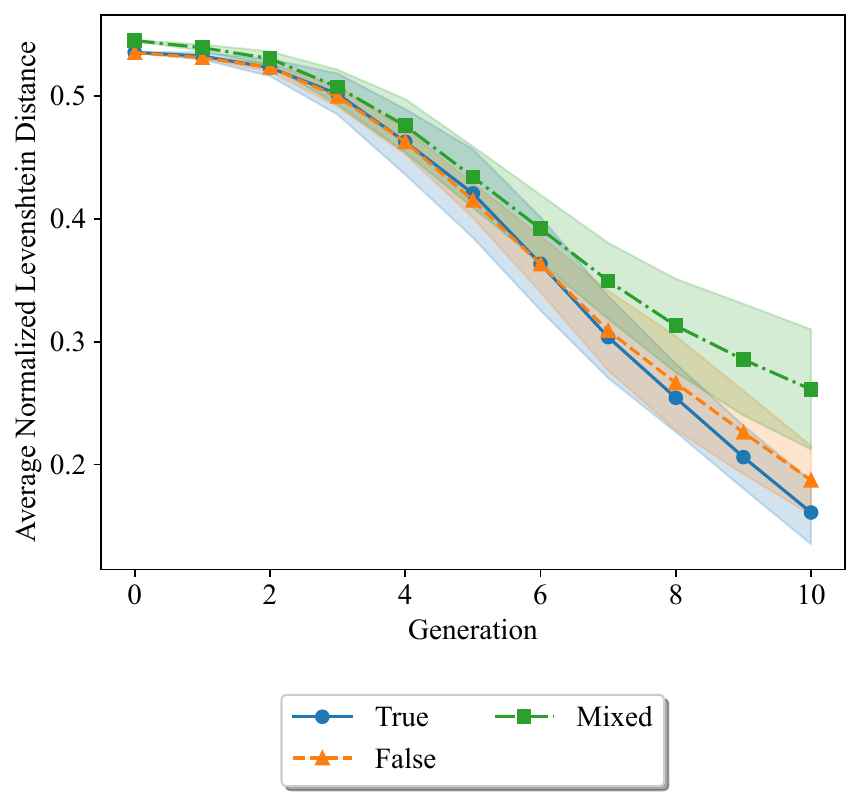}}
\caption{Average pairwise normalized Levenshtein distance over generations of a self-consuming training loop for the full synthetic data cycle with different initializations. Displayed is the mean and standard deviation at each generation for each initialization over 5 runs.}
\label{fig:diversity_per_init}
\end{center}
\vskip -0.2in
\end{figure}

So far we only looked at the self-consuming training loop starting from an initial dataset  consisting only of \texttt{True} expressions. To check whether our findings also hold for different initializations of data across multiple runs, we conduct further experiments. Specifically, we analyze the full synthetic data cycle with the original dataset $\mathcal{D}_0$ only consisting of (1) \texttt{True} expressions, (2) \texttt{False} expressions, and (3) an equal mixture of both. 
The results from Sect.~\ref{sec:results} indicate that the effect of the self-consuming training loop for the full synthetic data cycle are already clearly visible in the first 10 generations. 
Therefore, we conduct experiments with only 10 generations. With this saved computing time, we conduct 5 independent runs for each initialization.

Figure~\ref{fig:quality_per_init} displays the composition of samples $\mathcal{S}_t$ generated during a self-consuming training loop. Each subplot presents the composition for a different initialization for the full synthetic data cycle averaged over 5 independent runs: a) only \texttt{True} expressions, b) only \texttt{False} expressions, and c) half \texttt{True} expressions and half \texttt{False} expressions. Every bar in a subplot displays the composition of $\mathcal{S}_t$ generated from model $\mathcal{M}_t$ at generation $t$, except for the first bar in each subplot which displays $\mathcal{D}_0$. The green portion of a bar indicates the number of syntactically correct expressions that evaluate to \texttt{True} in that sample. The yellow part of a bar represents the number of syntactically correct expressions that evaluate to \texttt{False}, and the red part of a bar displays the number of syntactically incorrect expressions that result in an error when being parsed.

Similar to our results in Sect.~\ref{subsec:quality}, we observe an initial drop of \texttt{True} expressions by 20\% in favor of \texttt{False} expressions for the initialization with only \texttt{True} expressions in the first generation. Afterwards, the number of \texttt{True} expressions in a sample increases again over the course of generations. 
For the initialization with only \texttt{False} expressions, we observe a similar effect, only in the opposite direction. The number of \texttt{False} expression first drops by around 20\% and then increases again until the whole sample consists of only \texttt{False} expressions.
When initialized with an equal amount of \texttt{True} and \texttt{False} expressions, we can observe that the proportion of both expression types first stays stable and then by generation $t=6$ starts to slightly shift towards \texttt{True} expressions. Upon further inspection of the 5 runs, we find that one run quickly shifts towards only \texttt{True} expressions, one run shifts quickly to only \texttt{False} expression, one run shifts slowly to \texttt{True} expressions, and two runs maintain equal proportions between \texttt{True} and \texttt{False} expressions across the 10 generations. 

Figure~\ref{fig:diversity_per_init} displays the diversity measured in the average pairwise normalized Levenshtein distance for the three different initializations. Displayed is the mean and standard deviation of five runs over the number of generations. We observe that the diversity declines for all three types of initialization, further confirming our results from Sect.~\ref{subsec:diversity}. The initializations with only \texttt{True} and only \texttt{False} expressions decline at around the same speed. The initialization with equal proportions of \texttt{True} and \texttt{False} expressions declines a bit slower but shows the same effect.

Overall these findings indicate that the effects of the self-consuming training loop apply to different initializations across multiple runs, partially mitigating the limitations mentioned in Sect.~\ref{sec:limitations}.

\end{document}